\pdfoutput=1

\documentclass[11pt]{article}

\usepackage[final]{acl}

\usepackage{times}
\usepackage{latexsym}
\usepackage{CJKutf8}
\usepackage{booktabs}
\usepackage{multirow}
\usepackage{array}
\usepackage[T1]{fontenc}
\usepackage[many]{tcolorbox}

\usepackage[utf8]{inputenc}

\usepackage{microtype}

\usepackage{inconsolata}

\usepackage{graphicx}
\newcommand{\dataset}{\textsc{Chengyu-Bench}}

\title{\dataset{}: Benchmarking Large Language Models for Chinese Idiom Understanding and Use}

\author{
  Yicheng Fu$^1$, Zhemin Huang$^1$, Liuxin Yang$^1$, Yumeng Lu$^1$, and Zhongdongming Dai$^2$ \\
  $^1$Stanford University, CA, USA \\
  $^2$University of California San Diego, CA, USA \\
  \{easonfu, zheminh, lyang822, yumenglu\}@stanford.edu$^1$ \\
  z1dai@ucsd.edu$^2$
}

\begin{document}
\begin{CJK*}{UTF8}{gbsn}
\maketitle
\begin{abstract}
Chinese idioms (成语, Chengyu) are concise four-character expressions steeped in history and culture, whose literal translations often fail to capture their full meaning. This complexity makes them challenging for language models to interpret and use correctly. Existing benchmarks focus on narrow tasks—multiple-choice cloze tests, isolated translation, or simple paraphrasing. We introduce \textbf{\dataset{}}, a comprehensive benchmark featuring three tasks: (1) \emph{Evaluative Connotation}, classifying idioms as positive or negative; (2) \emph{Appropriateness}, detecting incorrect idiom usage in context; and (3) \emph{Open Cloze}, filling blanks in longer passages without options. \dataset{} comprises 2,937 human-verified examples covering 1,765 common idioms sourced from diverse corpora. We evaluate leading LLMs and find they achieve over 95\% accuracy on \emph{Evaluative Connotation}, but only \textasciitilde85\% on \emph{Appropriateness} and \textasciitilde40\% top-1 accuracy on \emph{Open Cloze}. Error analysis reveals that most mistakes arise from fundamental misunderstandings of idiom meanings. \textbf{\dataset{}} demonstrates that while LLMs can reliably gauge idiom sentiment, they still struggle to grasp the cultural and contextual nuances essential for proper usage. The benchmark and source code are available at: \url{https://github.com/sofyc/ChengyuBench}.
\end{abstract}

\section{Introduction}

Rooted in stories, values, and traditions passed down through generations, Chinese idioms (成语, Chengyu) represent a rich part of the language and culture. Most idioms come from classical literature or ancient folklore, and summarize the essence of a story in a highly compact form~\cite{yang2006taxonomy}. Because of their simplicity and literary quality, idioms are highly prized in Chinese communication. They can elegantly convey complex ideas and show the speaker's thoughts.

Meanwhile, these properties make idioms challenging for computational models. Chinese idioms are non‑compositional and metaphorical. They usually follow the conventions of ancient Chinese, and depend on cultural and historical contexts for interpretation~\cite{chinese_idiom}. For large language models (LLMs), they learn from significant patterns and may lack the cultural grounding to understand idioms. Therefore, even state-of-the-art Chinese LLMs can misinterpret idioms~\cite{li2024translate}.

Despite the importance of Chinese idioms, existing NLP benchmarks handle them only peripherally. For instance, ChID~\cite{chid} provides a large-scale cloze-style reading comprehension task; ~\citet{chinese_idiom} collects 115K sentence pairs in which idiomatic sentences are translated into non-idiomatic sentences. Cloze tests and paraphrase tasks are widely used to assess language proficiency~\cite{jonz1991cloze, tremblay2011proficiency, tan2021learning}, but they are not sufficient for a thorough evaluation of Chinese idioms: cloze tests mainly assess idiom retrieval or simplification, while the paraphrase task only measures lexical similarity. Moreover, general Chinese benchmarks, such as CLiMP~\cite{xiang2021climp}, do not include specialized idiom tasks. In short, existing benchmarks either overlook idiomatic expressions or lack scenarios that reflect real-world usage.

\begin{figure*}[t]
    \centering
    \includegraphics[width=\linewidth]{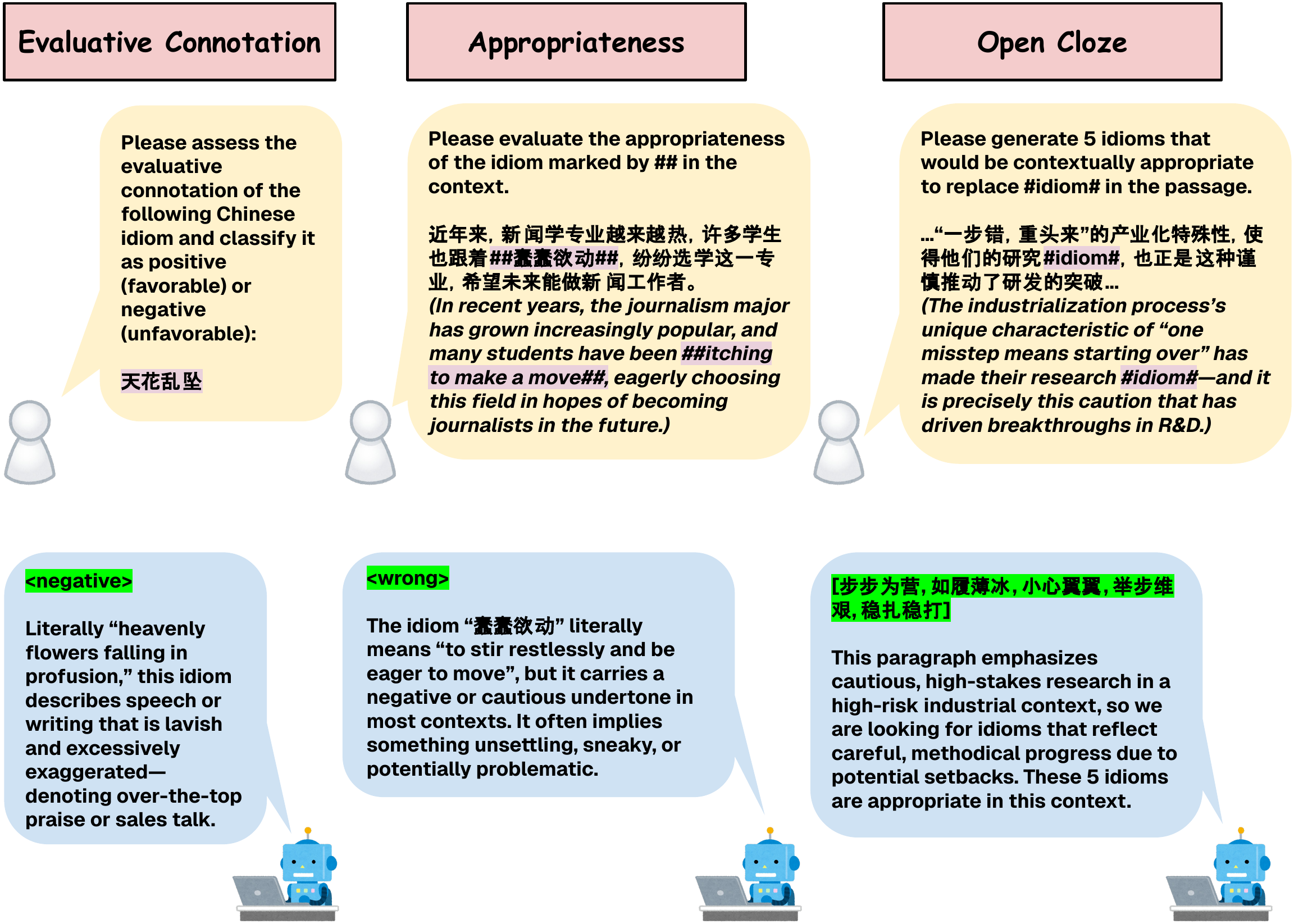}
    \caption{Subtask example. In the \textbf{Evaluative Connotation} subtask, the model must classify the sentiment polarity of a single idiom. In the \textbf{Appropriateness} subtask, it must decide whether the highlighted idiom fits the given context. In the \textbf{Open Cloze} subtask, it generates five idiom candidates, ranked by confidence, to complete the paragraph. Purple text highlights the idiom or placeholder in the prompt, and green text shows the answer extracted for evaluation.}
    \label{fig:task}
\end{figure*}

To mitigate the gap, we identified three core tasks that are lacking in existing benchmarks: evaluative connotation (categorizing the sentiment of idiomatic expressions in context), contextual appropriateness (determining whether candidate idioms are appropriate in context), and open cloze (generating idioms that are appropriate for the context in a given situation). These tasks reflect the actual requirements of real-world idiom usage. Figure~\ref{fig:task} shows some examples of these tasks. To the best of our knowledge, no current Chinese NLP benchmarks evaluates models across the full spectrum of idiom usage.

Our contributions are summarized as follows:
\begin{itemize}
\item We present \dataset{}, the first comprehensive benchmark for Chinese idiom understanding, built from diverse, naturally occurring texts. It includes over 3,000 human-annotated examples spanning 1,765 idioms, and features three tasks of increasing difficulty to holistically evaluate idiomatic proficiency.
\item We evaluate a wide range of state‐of‐the‐art LLMs and conduct error analysis on \dataset{}, discovering significant gaps between idiom recognition and proper usage.
\end{itemize}


\section{Related Work}

\subsection{Challenges of Chinese Idiom Understanding for LLMs}
Chinese idioms pose challenges to LLMs across semantic, structural, and cultural levels~\cite{chinese_idiom}. First, idioms' meanings often cannot be deduced from the constituent words. For example, "亡羊补牢" does not literally refer to mend the fence after sheep are lost, but rather implies that it is never  too late to try~\cite{chid}. This metaphorical nature requires models to understand the non-literal meaning. Second, idioms have a fixed structure, usually four characters, and cannot be decomposed and recomposed~\cite{kang2022study}. Third, idioms contain rich cultural and historical knowledge~\cite{chinese_idiom}. Many of them derive from classical literature or ancient anecdotes. Therefore, understanding Chinese idioms requires a deep understanding of Chinese tradition and history. Finally, the meanings and usages of idioms are highly context-dependent. Many idioms also have closely related counterparts, but with subtle differences in meaning or usage, making them more difficult to select or interpret~\cite{chid, chinese_idiom}.

These characteristics make idioms a rigorous testing ground for LLMs, which often demonstrate substantially lower proficiency in idiom-related tasks compared to human performance~\cite{chid, wu2024mitigating}.

\begin{table*}[h] 
\centering 
\begin{tabular}{>{\raggedright\arraybackslash}m{1.6cm} >{\raggedright\arraybackslash}m{3.2cm} >{\raggedright\arraybackslash}m{6.6cm} >{\raggedright\arraybackslash}m{2cm}}
\toprule 
\textbf{Idiom} & \textbf{Surface Meaning} & \textbf{True Meaning} & \textbf{Polarity} \\
\midrule
弹冠相庆 & Brushing off hats and celebrating together & Celebrating preemptively because they expect to gain advantages through improper means like cronyism or corruption & Negative \\ 
\midrule
舞文弄墨 & Waving writings and playing with ink & Using writing skills in a petty, deceptive, or manipulative way rather than for something noble or constructive & Negative \\
\midrule
惨淡经营 & Managing operations under miserable and bleak conditions & Persistently struggling and carefully managing things through hardship and difficulty, often with little reward & Positive \\
\bottomrule 
\end{tabular}  
\caption{Surface meaning, true meaning, and polarity of example Chinese idioms.} 
\label{tab:polarity} 
\end{table*}  

\subsection{Chinese Idiom Dataset}
The ChID dataset~\cite{chid} is a large-scale cloze test dataset, containing 581k passages and
729k blanks from three domains (news, novels, and essays). Each blank is accompanied by several candidate idioms, requiring models to select the most appropriate idiom. This dataset has become the standard benchmark for evaluating Chinese idiom comprehension~\cite{xu2020clue}. The CIP dataset~\cite{chinese_idiom} contains 115k sentence pairs. In each pair, one sentence contains a specific Chinese idiom while the other paraphrases its meaning in plain language. IdiomKB~\cite{li2024translate} includes 8,643 idiom interpretations in Chinese, English, and Japanese, evaluating models' idiom comprehension and translation abilities. However, these datasets primarily focus on limited tasks: cloze tests, paraphrasing, and translation, which cannot thoroughly determine whether idioms are being used appropriately in wider contexts.

\subsection{General Chinese Benchmarks for LLMs}
CLUE~\cite{xu2020clue} is the first large-scale benchmark for Chinese language understanding, consisting of nine sub-tasks, including semantic matching, short and long text classification, and reading comprehension, etc. C-Eval~\cite{huang2023ceval} focuses on higher-order knowledge and reasoning skills. It consists of 13,948 multiple-choice questions spanning 52 subjects, including science, engineering, humanities and social sciences. Inspired by the English MMLU benchmark~\cite{mmlu}, CMMLU~\cite{li2023cmmlu} is a comprehensive multitask Chinese benchmark covering 67 Chinese topics. More recently, WenMind~\cite{cao2024wenmind} is a comprehensive benchmark for Chinese Classical Literature and Language Arts (CCLLA). Although some general benchmarks~\cite{cao2024wenmind} include idiom-related subtasks, e.g. idiom explanation, the scale and diversity of these subtasks remain limited.

\begin{table*}[h]
\centering
\begin{tabular}{>{\raggedright\arraybackslash}m{2.4cm} >{\raggedright\arraybackslash}m{6.6cm} >{\raggedright\arraybackslash}m{5.2cm}}
\toprule
\textbf{Misuse Type} & \textbf{Example Idiom and Incorrect Usage} & \textbf{Explanation} \\
\midrule
Wrong Polarity & 
他在事故中失去了家人，但我们祝他\#\#一帆风顺\#\#。\newline
\textit{He lost his family in an accident, but we wish him \#\#smooth sailing\#\#.}
& Using a highly positive idiom in a tragic or sad situation \\
\midrule
Wrong Subject/Object & 
这台洗衣机\#\#毛遂自荐\#\#，功能强大。\newline
\textit{This washing machine \#\#volunteered itself\#\# and has great functions.}
& Idioms about human actions wrongly applied to objects \\
\midrule
Literal Misinterpretation & 
他把那只鹿说成是马，真是\#\#指鹿为马\#\#的好例子。\newline
\textit{He called that deer a horse, what a good example of \#\#calling a deer a horse\#\#.}
& Taking the idiom literally without understanding its deeper political or metaphorical meaning \\
\midrule
Incorrect Degree & 
他今天买了一杯咖啡，真是\#\#惊天动地\#\#的大事。\newline
\textit{He bought a cup of coffee today, what an \#\#earth-shaking\#\# event.}
& Using a highly exaggerated idiom for a trivial action or event \\
\bottomrule
\end{tabular}
\caption{Common misuse types of Chinese idioms with incorrect examples and explanations.}
\label{tab:misuse}
\end{table*}

\begin{table*}[h]
\centering
\resizebox{\textwidth}{!}{%
\begin{tabular}{>{\raggedright\arraybackslash}m{2.5cm} >{\raggedright\arraybackslash}m{7.2cm} >{\raggedright\arraybackslash}m{2.2cm} >{\raggedright\arraybackslash}m{2.2cm} >{\raggedright\arraybackslash}m{6cm}}
\toprule
\textbf{Task} & \textbf{Example} & \textbf{Available Options} & \textbf{Answer} & \textbf{Reason} \\
\midrule
\textbf{Evaluative Connotation} & 
好为人师 \newline
\textit{Fond of acting as a teacher to others.} & 
Positive, Negative & 
Negative & 
"好为人师" is used with the connotation of being overly eager to instruct others or assuming a superior attitude. \\
\midrule
\textbf{Appropriateness} & 
这次商品博览会，聚集了全国各地各种各样的新产品，真可谓\#\#浩如烟海\#\#，应有尽有。 \newline
\textit{This product expo gathered all kinds of new products from across the country; it can truly be said to be \#\#as vast as a sea of smoke\#\#, with everything one could possibly want.} & 
Correct, Wrong & 
Wrong & 
"浩如烟海" is used to describe the sheer quantity of writings, books, or documents. It emphasizes the overwhelming amount of texts, and cannot be used to describe physical goods. \\
\midrule
\textbf{Open Cloze} & 
...到达荒岛后，两人开始了他们的探险。起初，一切都显得那么平静和美好。然而，\#\#idiom\#\#，第三天晚上，他们遭遇了一群野兽的袭击。在混乱中，他们的干粮被野兽抢走，指南针也丢失了。 \newline
\textit{...Upon arriving at the island, the two began their exploration. At first, everything seemed so peaceful and wonderful. However, \#\#idiom\#\#, on the third night, they were attacked by a group of wild beasts. In the chaos, their dry food was stolen by the beasts, and their compass was lost.} & 
--- & 
好景不长 \newline
\textit{Good times do not last long} & 
The sentence transition requires an idiom that hints at a short-lived good situation turning bad, and "好景不长" exactly conveys this. \\
\bottomrule
\end{tabular}%
}
\caption{Examples, available options, correct answers, and reasoning for each subtask.}
\label{tab:example}
\end{table*}

\begin{figure*}
    \centering
    \includegraphics[width=1\linewidth]{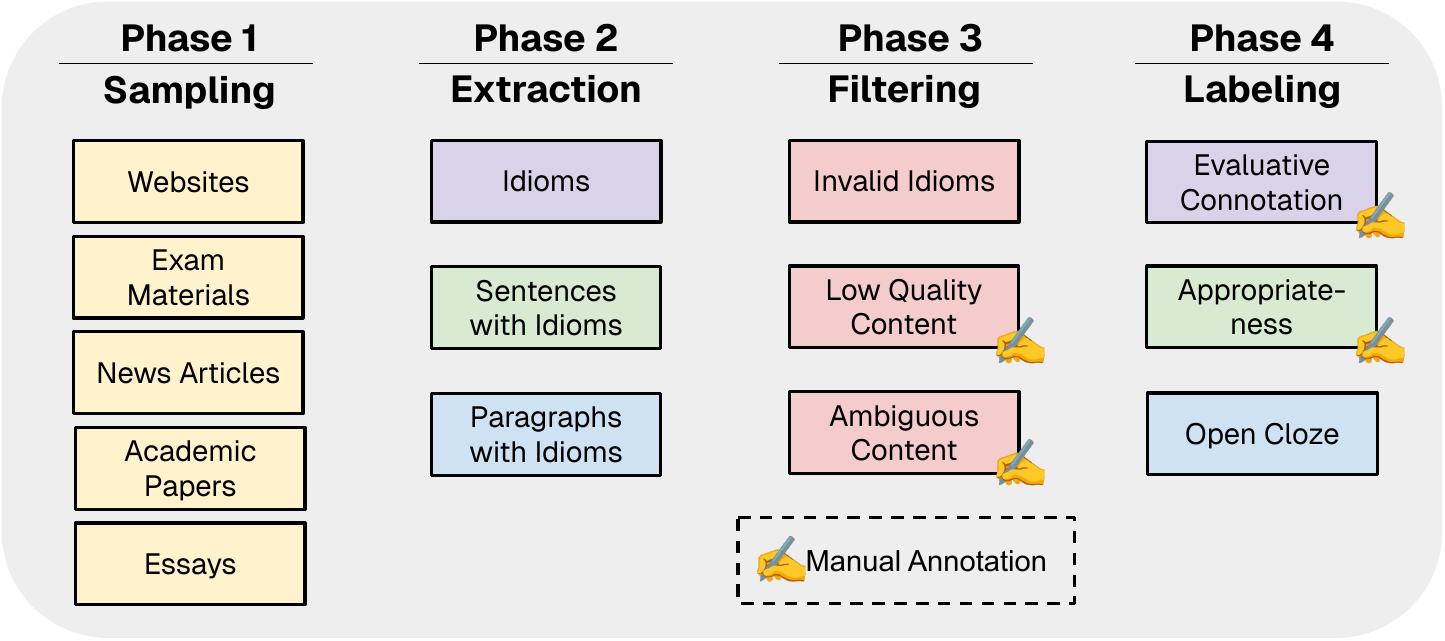}
    \caption{Overview of the benchmark generation pipeline. The process consists of four phases: (1) Sampling diverse high-quality sources, (2) Extracting idioms, sentences, and paragraphs, (3) Filtering invalid, low quality, or ambiguous content, and (4) Labeling data for polarity, appropriateness, and cloze tasks. Manual annotation is required during filtering and labeling stages.}
    \label{fig:pipeline}
\end{figure*}
\section{Benchmark}

\subsection{Task Definition}

Most existing Chinese idiom benchmarks are limited to narrow cloze tests—either choosing from a small set of options~\cite{chid} or completing very short sentences~\cite{jiang2018chengyu}. Others ask models to select an idiom based on its definition~\cite{wu2024mitigating} or to paraphrase sentences using idioms~\cite{chinese_idiom}. Yet none of these tasks fully assesses a model’s ability to understand and use idioms in realistic, extended contexts. To bridge this gap, we introduce three complementary subtasks: (1) identifying whether an idiom conveys a positive or negative sentiment \textbf{(Evaluative Connotation)}, (2) determining if an idiom is appropriately used in a sentence \textbf{(Appropriateness)}, and (3) filling in blanks with suitable idioms in long paragraphs \textbf{(Open Cloze)}. Detailed prompts for each subtask are provided in Appendix~\ref{appendix:prompts}.

\paragraph{Evaluative Connotation}
Chinese idioms often carry rich, culturally rooted sentiments that are not obvious from their literal wording. Table~\ref{tab:polarity} shows examples where surface meaning can mislead. Accurately identifying an idiom’s polarity—positive or negative—is essential for using it correctly in real‐world text. In this subtask, we challenge models to label each idiom’s sentiment polarity as conveyed by the writer.

\paragraph{Appropriateness}

Whether a Chinese idiom is correctly used in a sentence depends on multiple factors. One key factor is using an idiom with the correct polarity, as discussed earlier. Other common mistakes include choosing the wrong subject or object, misinterpreting the idiom literally, or applying an idiom with an inappropriate degree of intensity. Examples of these errors are shown in Table~\ref{tab:misuse}. Such misuse is very common among human writers, not to mention language models. Therefore, this task effectively tests whether a model can detect inappropriate idiom usage in Chinese sentences.

\paragraph{Open Cloze}
In this subtask, models must fill a blank in a longer passage without any provided options. We source these passages from online texts and ask each model to generate its top five idiom candidates, ranked by confidence. Allowing multiple predictions reflects real-world writing practices—authors often consider several idiomatic expressions before selecting the most appropriate one. This approach also accounts for the fact that multiple idioms may convey similar nuances; however, it is rare for more than five idioms to express the same meaning, as overly redundant expressions tend to fall out of use over time. This setup tests a model’s ability to recall and apply idioms unaided. 

Table~\ref{tab:example} illustrates example instances, their annotations, and the rationale behind the correct answer for each subtask.

\subsection{Benchmark Generation}

The overall pipeline for benchmark generation is shown in Figure~\ref{fig:pipeline}. It consists of four main steps:

\paragraph{Sampling} In this stage, we collect a corpus from diverse yet high-quality sources, including webpages, exam materials, news articles, academic papers, and essays. These materials are used as the foundation for constructing our benchmark.

\paragraph{Extraction} We extract three types of content from the corpus: individual idioms, sentences with idioms, and paragraphs with idioms. For the idiom vocabulary, we start with the 31,648 idioms listed in the official Xinhua Dictionary~\footnote{\url{https://github.com/pwxcoo/chinese-xinhua}}. Since many of these idioms are rarely used and provide limited practical value, we further filter them based on document frequency computed from online resources~\cite{han2016thuocl}, resulting in a final vocabulary of 7,208 commonly used idioms. All extracted content must contain idioms from this filtered vocabulary.

For sentences and paragraphs, we prioritize extracting paragraphs whenever multiple sentences are available.
If only a single sentence is available—which is often the case in exam materials—we extract the sentence directly.

\paragraph{Filtering} Some filtering is already performed during extraction, such as removing invalid or low-frequency idioms. In addition, we manually filter out low-quality content, such as webpages that simply list idioms without context, or ambiguous content, such as cases where an idiom's meaning has recently changed or is controversial.

\paragraph{Labeling} In the final stage, we annotate the data according to each subtask. For individual idioms (Evaluative Connotation), we keep only those with an unambiguous positive or negative sentiment, manually discarding neutral cases to avoid confusion. For sentences containing idioms (Appropriateness), we label each example as correctly or incorrectly used—most correct instances come from online corpora, while negative examples are drawn from exam materials and educational sites that train students to spot misuse. For paragraphs with idioms (Open Cloze), we replace the target idiom with a placeholder (\#\#idiom\#\#) for the model to predict. If a sentence or paragraph contains multiple idioms, we duplicate the example so that each idiom is treated as a separate data point.

\subsection{Benchmark Statistics}

To assess the quality of our benchmark, we conducted a detailed analysis focusing on the number of unique idioms and the average document frequency of idioms from online resources across each subtask. Table~\ref{tab:number} summarizes the number of data points and unique idioms for each task. Notably, in the Evaluative Connotation task, each data point corresponds to a unique idiom, which aligns with the task design where each entry is centered on a single idiom. In total, our dataset covers 1,765 unique idioms, with an average of approximately 1.66 data points per idiom.

\begin{table}[h] 
\centering 
\resizebox{\columnwidth}{!}{%
\begin{tabular}{lcc} 
\toprule 
\textbf{Task Category} & \textbf{\# of Data Points} & \textbf{\# of Unique Idioms} \\
\midrule
\textbf{Connotation} & 540 & 540 \\
\textbf{Appropriateness} & 572 & 441 \\ 
\textbf{Open Cloze} & 1,825 & 1,067 \\ 
\midrule 
\textbf{Overall} & 2,937 & 1,765 \\
\bottomrule 
\end{tabular} 
} 
\caption{Number of data points and unique idioms across different subtasks in our benchmark.} 
\label{tab:number} 
\end{table}

To evaluate the representativeness of the idioms selected in our benchmark, we first found a comprehensive vocabulary of Chinese idioms with their document frequencies based on online resources \cite{han2016thuocl}. In this vocabulary, idiom frequencies range from a minimum of 21 to a maximum of 54,113, with an average frequency of 1,276. We then extracted the idioms appearing in each benchmark subtask and computed their average document frequency. As shown in Table~\ref{tab:frequency}, the idioms used in our benchmark have significantly higher average frequencies than those in the general vocabulary, suggesting that our dataset predominantly covers idioms that are commonly used in real-world Chinese language contexts.

\begin{table}[h] 
\centering 
\resizebox{\columnwidth}{!}{%
\begin{tabular}{lc} 
\toprule 
\textbf{Statistic / Task Category} & \textbf{Avg Document Frequency} \\ 
\midrule 
\textbf{Vocabulary Minimum (Min)} & 21 \\ 
\textbf{Vocabulary Maximum (Max)} & 54,113 \\ 
\textbf{Vocabulary Average (Avg)} & 1,276 \\ 
\midrule 
\textbf{Connotation} & 2,136 \\ 
\textbf{Appropriateness} & 2,890 \\ 
\textbf{Open Cloze} & 7,411 \\ 
\midrule 
\textbf{Overall} & 5,650 \\ 
\bottomrule 
\end{tabular}%
} 
\caption{Average document frequencies of idioms used in our benchmark compared to the general idiom vocabulary.} \label{tab:frequency} 
\end{table}

Table~\ref{tab:length} presents the average context token length for reading comprehension tasks in previous datasets and our \dataset{}. For translation and paraphrase tasks, we measure the length of the source sentences from the test split. For cloze test and appropriateness tasks, we measure the length of the given sentences or paragraphs.

We observe that the appropriateness task in ChengyuBench has an even longer average context length than the earlier cloze test dataset CCT~\cite{jiang2018chengyu}, demonstrating the increased complexity of the task.
Moreover, the cloze test in ChengyuBench is nearly three times longer than the previous cloze benchmark ChID~\cite{chid}, further highlighting the richness and difficulty of our dataset. To the best of our knowledge, this is the longest and most challenging Chinese idiom cloze test constructed to date.

\begin{table}[h] 
\centering 
\resizebox{\columnwidth}{!}{%
\begin{tabular}{llc} 
\toprule 
\textbf{Dataset} & \textbf{Task} & \textbf{Avg. Context Tokens} \\ 
\midrule 
\textbf{CIBB}~\cite{shao2017evaluating} & Translation & 23.13 \\
\textbf{CIP}~\cite{chinese_idiom} & Paraphrase & 43.75 \\
\textbf{CCT}~\cite{jiang2018chengyu} & Open Cloze & 54.72 \\
\textbf{ChID}~\cite{chid} & MC Cloze & 212.10 \\
\midrule
\multirow{2}{*}{\textbf{\dataset{}}} & Appropriateness & 56.91 \\ 
& Open Cloze & 600.41 \\ 
\bottomrule 
\end{tabular}%
}
\caption{Average context token length for Chinese idiom reading comprehension tasks. Our benchmark exhibits the longest contexts, highlighting its elevated difficulty.}
\label{tab:length} 
\end{table}

\section{Results}

Table \ref{tab:results} reports the complete performance of all evaluated LLMs on both our benchmark and the ChID dataset. In our experiments, we benchmark 5 closed-source models: Gemini-2.0-Flash, Gemini-2.5-pro~\cite{geminiteam2025}, Claude-3.7-Sonnet~\cite{anthropic2024claude} , GPT-4o~\cite{hurst2024gpt}, GPT-4.1 and 3 open-source models: DeepSeek-R1~\cite{guo2025deepseek}, DeepSeek-V3~\cite{deepseekv3} and Qwen2.5-72B~\cite{qwen25}.

\begin{table*}[h]
\centering
\resizebox{\textwidth}{!}{%
\begin{tabular}{lcc|cccc|c}
\toprule
\multirow{2}{*}{\textbf{Model}} & 
\multirow{2}{*}{\textbf{Connotation}} & 
\multirow{2}{*}{\textbf{Appropriateness}} & 
\multicolumn{4}{c|}{\textbf{Open Cloze}} & 
\multirow{2}{*}{\textbf{ChID Acc.}} \\
\cmidrule(lr){4-7}
& & & \textbf{Acc.@1} & \textbf{Acc.@3} & \textbf{Acc.@5} & \textbf{Valid Idiom} & \\
\midrule
Random      & 50.00 & 50.00 & --- & --- & --- & --- & 14.29 \\
\midrule
\multicolumn{8}{l}{\textbf{Closed-Source Models}} \\

\midrule
Gemini-2.0-Flash      & 95.19 & 55.07 & 15.01 & 27.18 & 30.85 & \textbf{86.65} & 56.00 \\
Gemini-2.5-Pro        & 97.04 & 73.95 & \textbf{40.05} & \textbf{55.40} & \textbf{60.77} & 73.10 & \textbf{75.60} \\
Claude-3.7-Sonnet     & 95.19 & 61.89 & 23.78 & 37.37 & 42.30 & 67.77 & 64.20 \\
GPT-4o                & 96.11 & 71.15 & 18.19 & 28.16 & 31.95 & 69.75 & 59.65 \\
GPT-4.1               & 97.04 & 66.26 & 23.51 & 35.51 & 39.34 & 66.68 & 63.35 \\
\midrule
\multicolumn{8}{l}{\textbf{Open-Source Models}} \\
\midrule
DeepSeek-R1           & \textbf{97.56} & \textbf{83.27} & 27.12 & 38.05 & 42.23 & 80.73 & 72.80 \\
Qwen2.5-72B           & 95.74 & 56.64 & 24.99 & 33.37 & 36.77 & 71.65 & 65.80 \\
DeepSeek-V3           & 97.22 & 74.83 & 33.59 & 45.75 & 48.99 & 82.10 & 69.30 \\
\bottomrule
\end{tabular}
}
\caption{Comprehensive performance (\%) of different models on the Evaluative Connotation, Appropriateness, and Open Cloze subtasks of our benchmark, as well as accuracy on the ChID dataset. Acc.@k denotes the proportion of examples in which the correct idiom appears within the model’s top-k predictions; Valid Idiom indicates the percentage of predicted idioms that are listed in the Xinhua Dictionary.}
\label{tab:results}
\end{table*}

\paragraph{Performance Gap Between Connotation and Other Subtasks}  
All models achieve over 95\% accuracy on Evaluative Connotation, indicating that modern LLMs reliably grasp basic sentiment polarity of Chinese idioms. In contrast, Appropriateness scores drop below 85\%, and Open Cloze accuracy@1 falls to 40\% or lower. This widening gap underscores that while sentiment recognition is effectively mastered, understanding contextual and cultural nuances to correctly use idioms remains challenging.

\paragraph{Model Comparison}  
Among all LLMs, Gemini-2.5-Pro leads across all Cloze metrics and also attains the highest ChID accuracy. DeepSeek-R1 excels at Appropriateness (83.27\%) and Evaluative Connotation (97.56\%), reflecting its strong contextual understanding. DeepSeek-V3 delivers the most balanced profile, with competitive Appropriateness and a high Valid Idiom rate, even outperforming its reasoning-focused variant in Open Cloze. Interestingly, Gemini-2.0-Flash yields the best Valid Idiom ratio (86.65\%) despite lower overall task performance, suggesting that over-reliance on dictionary validity does not guarantee correct usage.

\paragraph{Performance of Chinese LLMs}  
China-developed models in the DeepSeek series show distinct advantages. Both DeepSeek-R1 and DeepSeek-V3 outperform most others in Appropriateness and Valid Idiom rate, indicating superior capture of cultural and contextual signals essential for idiom usage. Their strong results likely stem from specialized training on richer Chinese corpora and tailored optimizations for native linguistic patterns.

\begin{table*}[h]
\centering
\resizebox{\textwidth}{!}{%
\begin{tabular}{>{\raggedright\arraybackslash}m{2.5cm} >{\raggedright\arraybackslash}m{6.2cm} >{\raggedright\arraybackslash}m{7.0cm}}
\toprule
\textbf{Error Type} & \textbf{Definition} & \textbf{Example} \\
\midrule
Meaning Misinterpretation &
The model misunderstands an idiom’s core semantics and so mislabels correct uses as incorrect (or vice versa). &
It reads "山高水低" (\textit{mountains high, waters low}) as strictly about fatal mishaps, whereas the benchmark treats it as an acceptable metaphor for any looming hardship. \\
\midrule

Domain Adaptation Error &
The model fails to transfer an idiom from its original domain into a new context, rejecting valid extensions. &
It treats "师出无名" (\textit{army sent without a name}) as only military jargon and flags its bureaucratic sense ("no justification for approval") as wrong. \\
\midrule

Collocation \& Register Oversight &
The model ignores whether a perfectly grammatical but uncommon collocation is acceptable, or whether register shifts are fine. &
It marks "林林总总" (\textit{numerous and varied}) wrong simply because "林林总总" more often describes things, not book characters in this context. \\
\midrule

Connotation Polarity Confusion &
The model mixes up an idiom’s positive/neutral vs. negative undertone. &
It judges "心照不宣" (\textit{implicit mutual understanding}) as collusive wrongdoing when the benchmark counts it as a neutral implicit agreement. \\
\midrule

Presupposition Ignorance &
The model overlooks built-in requirements of an idiom—like needing a mix of good/bad  or a sharp qualitative contrast—and so misfires. &
It labels "泥沙俱下" (\textit{sand and silt flow together}) wrong because it sees only negative examples, even though the benchmark permits it in contexts of mixed quality. \\
\bottomrule
\end{tabular}%
}
\caption{Common error types and corresponding examples in idiom-appropriateness classification.}
\label{tab:error-types}
\end{table*}

\subsection{Error Analysis of the Appropriateness Task}

\begin{figure}
    \centering
    \includegraphics[width=\linewidth]{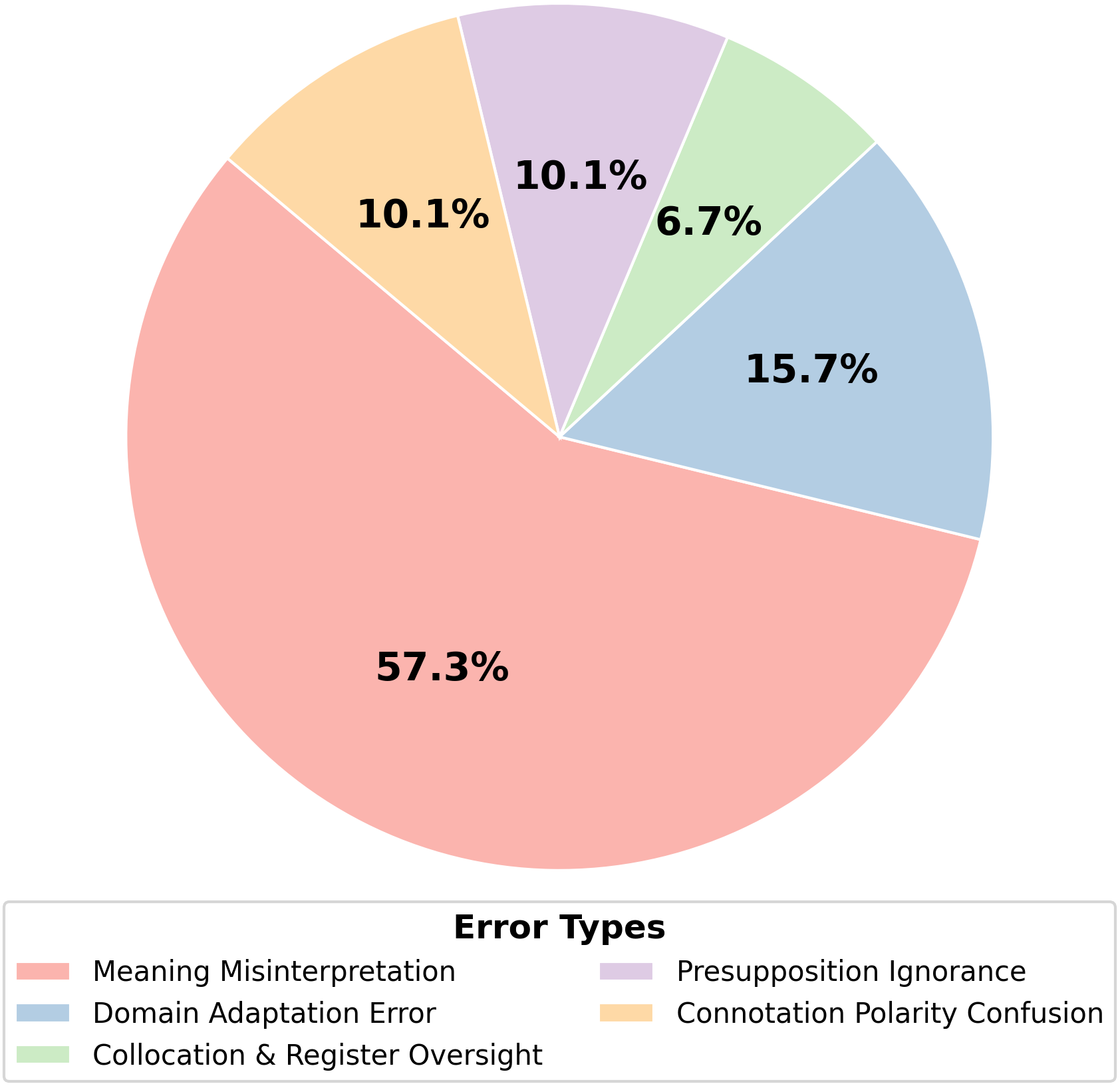}
    \caption{Distribution of error types made by Deepseek-R1 on the idiom appropriateness task.}
    \label{fig:error-distribution}
\end{figure}

To investigate why the model errs on the Chinese idiom appropriateness task, we conducted a detailed error analysis. First, we grouped the possible mistakes into five categories (see Table \ref{tab:error-types}), spanning from basic meaning misinterpretation to failures in context comprehension, usage adaptation, and connotation polarity. Next, we asked Gemini 2.5 Pro to label each error made by our best-performing LLM, Deepseek-R1, according to its reasoning trajectory. Figure \ref{fig:error-distribution} shows the resulting distribution of error types. Meaning misinterpretation is by far the most frequent, accounting for 57.3\% of all errors. This is followed by domain adaptation errors, where the model understands the idiom’s literal meaning but fails to apply it correctly in a new context. Collocation and register oversight appears least often. Overall, these findings suggest that—even at its best—current LLMs still struggle with fundamental idiom understanding, and have yet to master more advanced reasoning.

\section{Conclusion}
In this work, we introduce \dataset{}, a comprehensive benchmark designed to evaluate LLMs' understanding and usage of Chinese idioms across three distinct tasks: evaluative connotation, contextual appropriateness, and open cloze completion. Our benchmark addresses significant gaps in existing Chinese idiom evaluation datasets by providing longer and context-rich examples that more accurately reflect real-world language usage.

Our experimental results reveal a disparity between models' performance on different tasks. While contemporary LLMs demonstrate strong performance on identifying the evaluative connotation of idioms, they struggle considerably with determining appropriate usage and perform even more poorly on generating suitable idioms in context. This performance gap highlights that understanding sentiment does not guarantee mastery of the cultural nuances needed for proper idiom usage. Error analysis further reveals that the majority of mistakes stem from basic meaning misinterpretation, suggesting that even leading models still struggle with the fundamental semantics of Chinese idioms. 


\dataset{} provides a rigorous testing ground for evaluating culturally-specific language understanding in LLMs. We hope this work will inspire future research on idiom comprehension, advancing AI systems with deeper understanding of linguistic and cultural nuances in Chinese and potentially other languages.

\clearpage

\section*{Limitations}
While \dataset{} is the most comprehensive idiom task dataset to our knowledge and yields clear empirical insights into how contemporary LLMs handle real-world idiom use, several factors naturally delimit our study and also suggest where the benchmark can evolve.

Our benchmark focuses exclusively on canonical four-character chengyu and, in the Evaluative Connotation task, employs a binary polarity scheme; thus, longer proverb forms, context-dependent sentiment shifts, and emerging internet idioms fall outside the current scope.

Moreover, although we evaluate the most common idioms usage: recognition, misuse detection, and generative insertion, other minor idiom-oriented skills—such as paraphrasing, cross-lingual translation, and analogy—remain unexplored.

Also, it is worth noting that LLMs are increasingly deployed as components of compound AI systems—e.g., LLM agents \cite{li2024personal, fu2024camphor} or retrieval-augmented generation (RAG) architectures \cite{lewis2020retrieval, fu2025conquer}. However, our benchmark focuses exclusively on standalone LLMs and does not cover these more complex configurations.

Lastly, we anticipate regular updates on the dataset, since idiom popularity and nuance shift with cultural discourse, and advances in prompting strategies and LLM capabilities will continue to refine performance estimates.



\nocite{*}
\bibliography{main}

\clearpage
\appendix

\section{Prompts}
\label{appendix:prompts}

Here is the prompt for the Evaluative Connotation subtask:
\begin{tcolorbox}[colframe=black, colback=gray!10, title=Evaluative Connotation, breakable]
Please determine the evaluative connotation of the following Chinese idiom. Classify the idiom as either positive (with a favorable meaning) or negative (with an unfavorable meaning). Do not choose neutral. \\

The idiom is as follows: \\
\{idiom\} \\

Please provide your final answer in the format: \\
<positive> or <negative>
\end{tcolorbox}

Here is the prompt for the Appropriateness subtask:
\begin{tcolorbox}[colframe=black, colback=gray!10, title=Appropriateness, breakable]
Below is a Chinese passage. Please evaluate the appropriateness of the idiom marked by \#\# within the given context. Determine whether the idiom is used correctly or incorrectly based on its meaning and usage in standard Chinese. \\

The passage is as follows: \\
\{sentence\} \\

Please provide your final answer in the format: \\
<correct> or <wrong>
\end{tcolorbox}

Here is the prompt for the Open Cloze subtask:
\begin{tcolorbox}[colframe=black, colback=gray!10, title=Open Cloze, breakable]
Below is a Chinese passage. Please generate five four-character idioms that would be contextually appropriate to replace the placeholder \#idiom\# in the passage. \\

The passage is as follows: \\
\{paragraph\} \\

Please rank the idioms from most to least appropriate based on the context. At the end of your response, provide the idioms in the following format between <answer> and </answer>: \\
<answer><idiom1, idiom2, idiom3, idiom4, idiom5></answer> \\
Do not output any additional content between <answer> and </answer>.
\end{tcolorbox}

Here is the prompt for error analysis for Appropriateness subtask:
\begin{tcolorbox}[colframe=black, colback=gray!10, title=Error Analysis for Appropriateness, breakable]
We're evaluating whether a model can correctly judge if the idiom marked by \#\# fits its context. Below you'll find an example where the model made a mistake in answer. Your task is to identify the single most likely error type for each case, choosing from the list provided. \\

Error Types: \\

1. Meaning Misinterpretation \\
The model misunderstands an idiom's core meaning, causing it to misjudge correct usage (or vice versa). \\

2. Domain Adaptation Error \\
The model fails to apply an idiom correctly when it appears in a new or extended context. \\

3. Collocation \& Register Oversight \\
The model ignores whether a rare but valid collocation or an acceptable shift in formality is appropriate. \\

4. Connotation Polarity Confusion \\
The model confuses an idiom's positive, neutral, or negative tone. \\

5. Presupposition Ignorance \\
The model overlooks an idiom's inherent requirements—such as needing contrasting elements—and thus misclassifies usage. \\

Example: \\
Paragraph: \{paragraph\} \\
Correct Answer: \{label\} \\
Model Reasoning: \{reasoning\} \\
Model Answer: \{answer\} \\

Please pick one of the five error types above and output only its name:
\end{tcolorbox}

\end{CJK*}
\end{document}